# STRONG AND WEAK METHODS: A LOGICAL VIEW OF UNCERTAINTY

John Fox
Imperial Cancer Research Fund Laboratories, London

Before about 1660 when the modern Pascalian concept of probability emerged we had no formal methods for making predictions from incomplete or unreliable information (Hacking 1975). Since then many rival interpretations of probability have appeared and some have disappeared. Philosophical issues about the true meaning of probability remain (eg whether probability is a subjective or objective concept) but, nevertheless, a formidable range of practical numerical methods has been established for making predictions under uncertainty.

The appearance of Artificial Intelligence (AI) and expert systems are stimulating a new debate. AI systems of all kinds need to make decisions, often under uncertainty. Statistical decision theory has the concept of mathematical probability to represent uncertainty but AI emphasises "heuristics". Does AI imply new views of uncertainty, or only the adaptation of standard methods?

Practical developments suggest the latter. So far expert system designers have only modified deterministic knowledge representations like semantic nets and production rules with numerical belief coefficients (Fox, 1985). Although the techniques may be formally criticised (Spiegelhalter, 1985) fuzzy logic, Bayesian methods and the Shafer-Dempster theory of belief functions have all been used with some success.

A more radical view is that AI's goals require a reorientation of ideas about probability. Classical statistics deliberately seeks models which are abstractions (simplifications) of the world, but models are only valid if features of the problem do not violate their assumptions. Intelligent problem solvers (eg autonomous robots, learning programs, and statisticians) encounter situations which are ill-defined; the validity of the assumptions is dubious or unknown. Under these circumstances we should either adapt a method to the idiosyncrasies of the situation or adopt methods for dealing with uncertainty which make weaker assumptions.

AI has always been especially concerned with ill-defined problems. It advocates "weak methods" for problem solving which embody only weak assumptions. To illustrate, the knowledge that "claiming a breakthrough in cancer is unwise" influences my organisation's public relations policy. This heuristic is hard to formalise (what is a breakthrough?) but not vacuous. In the present state of the art, formalisation would introduce unrealistic restrictions on its possible meanings.

Decision making under uncertainty is often an ill-defined problem. When new problems are encountered (eg by a doctor), or invented (eg by a scientist) little may be known about their structure. To date expert systems have mostly been used for routine decisions where the structure is relatively well known. For the future we need weak, knowledge based methods for situations where the assumptions of strong, formal methods are too fierce.

## Non-numerical methods as weak methods

One place where restrictive assumptions creep in is in the model of quantitative and distributional features of the application domain (eg conditional independence assumptions). Paul Cohen has explored non-numerical methods for coping with uncertainty in his Theory of Endorsements. He recommends evaluation of hypotheses through the reasons (endorsements) for believing or disbelieving them.

There are many insights in Cohen's work but I am not sure that it satisfies the need for weak methods. We noted two strategies for dealing with idiosyncratic problems. (1) Elaborate the model so it takes explicit account of the details or (2) use a method less sensitive to the details. Cohen's approach seems to be of the first kind. The ability to deal with reasons for beliefs may be useful but it entails risks that the method may "be too cumbersome to be useful" and may "make unrealistic assumptions about the accuracy or availability of the evidence it requires" (Cohen, p 186).

An alternative is to relax quantitative assumptions by substituting a qualitative (logical) vocabulary for talking about different kinds of uncertainty. I consider three types here; possibility, probability and plausibility, after the next section which places them in a general context.

## Uncertainty as a type of knowledge

Although knowledge is a general concept most expert systems treat it narrowly, dealing only with the facts, rules etc of a domain. A wider view is that the domain is only one source of knowledge. Some knowledge is general (eg about the behaviour of physical objects), some more limited (eg knowledge about the current problem). Uncertainty knowledge is panoramic; it is a specialised form of knowledge, yet it straddles all other types of general, domain and problem specific knowledge bases (Fox, 1975). This figure illustrates the relationship.

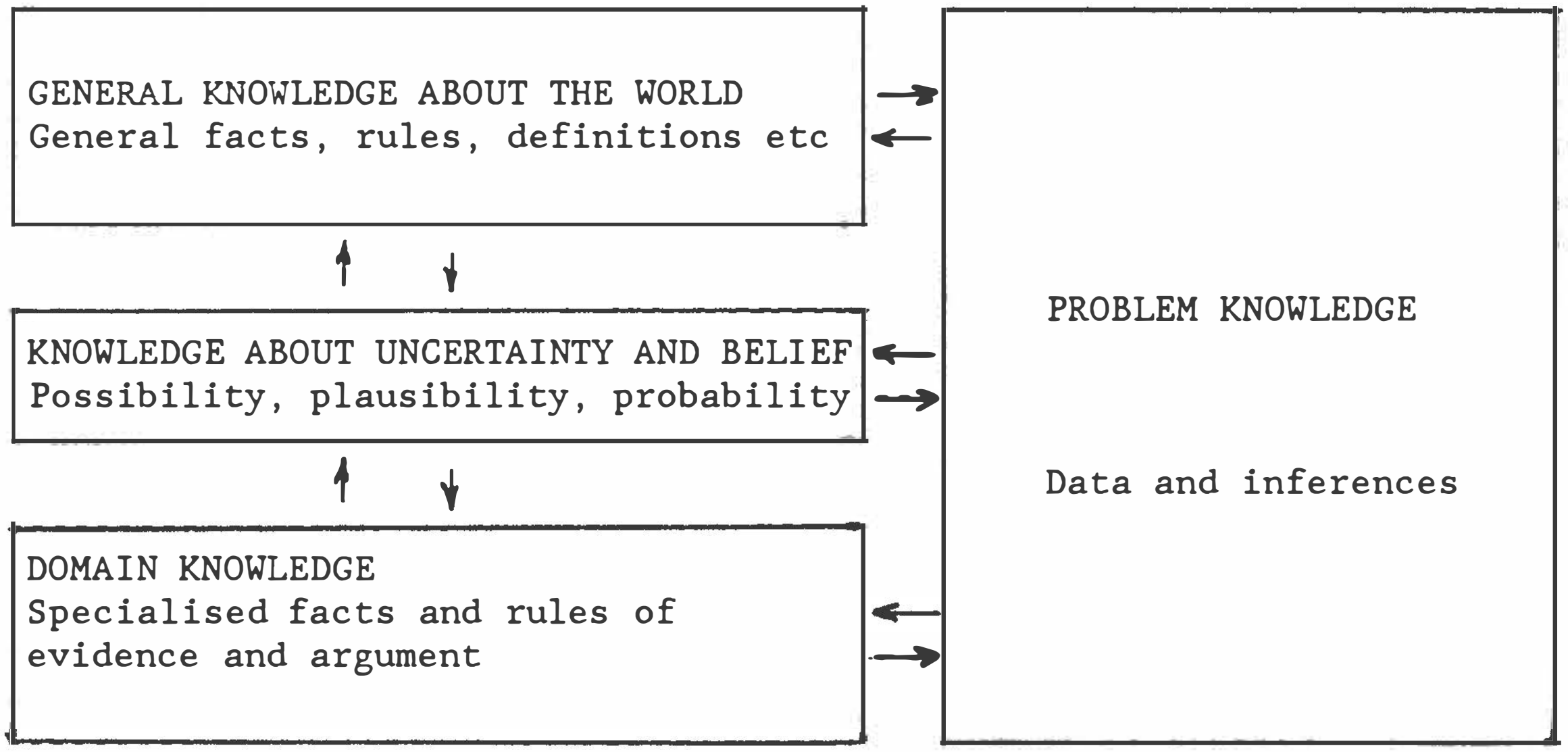

## Possibility, probability and plausibility

POSSIBILITY. Consider some proposition P, which may be any fragment of knowledge - a general rule, a domain fact or an item of data. P is said to be possible if no conditions that are necessary for P are violated. P may have no necessary conditions, or we may be ignorant of the state of these conditions; these circumstances do not affect a statement of possibility. We may also have evidence for or against P without affecting possibility. Possibilistic methods can introduce propositions into problem knowledge without commitment to a degree of belief. This is a fragment of possibility knowledge expressed as a logic program:

```
P is possible if not( P is not possible ).
P is not possible if P is exclusive of Q
                and Q is certain.
P is not possible if P is impossible.
P is impossible if C is a necessary condition for P
                and C is not possible.
P is not impossible if not( P is impossible ).
```

'Necessary conditions' are domain knowledge, or derived by applying general knowledge to domain knowledge.

PLAUSIBILITY. P is plausible if P is possible, and there is an argument in support of P or the arguments for P are stronger than those against (however we define 'stronger'). In "it is plausible that there are sub-atomic particles yet to be discovered" the plausibility argument might be a mathematical one from certain physical assumptions.

Where arguments exist for and against P (eg from different axioms) then we may use various strategies to decide plausibility. These include preferring arguments from general principles rather than special cases, or short arguments rather than long ones. P may be plausible whether or not there is evidence to support P, and even if there is evidence against. This is an illustrative logic program for plausibility:

```
P is plausible if A is argument for P and
                 not( P is exclusive of Q and
                      Q is more plausible than P ).
P is not plausible if not( P is plausible )
P is not plausible if P is implausible.
P is implausible if P is exclusive of Q and
                    Q is more plausible than P.
P is not implausible if not( P is implausible ).
```

Arguments include special case (P is a special case of P'), analogy (P is an analogy of P'), and model based arguments (P follows from some causal model M). Strategies of argument are domain independent, but particular special-general relationships, analogies, models etc may be domain specific. The relationship 'is more plausible than' requires a function to order its arguments which may or may not be numerical.

PROBABILITY. P is probable if P is possible and there is at least one item of evidence in favour of P. When evidence goes both ways we may again choose non-numerical strategies for resolving the conflict, eg to prefer direct evidence over indirect. This a logic program for probability:

```
P is probable if E is evidence for P and
                 not( P is exclusive of Q and
                      Q is more probable than P ).
P is not probable if not (P is probable).
P is not probable if P is improbable.
P is improbable if P is exclusive of Q
                 and Q is more probable than P.
P is not improbable if not (P is improbable ).
```

Evidence for P may be observations of P, reports of P, or signs associated with P (eg frequentistically). The relationship 'is more probable than' requires a function to order its arguments. This need not be numerical (eg show one set of evidence to be an experimental artifact) but numerical functions are best known, eg those which compute likelihood ratios.

## Revision of belief

As knowledge about a problem is obtained we may have to revise beliefs in hypotheses. This classical problem must be addressed by weak methods as well as strong ones. The problem of combining information is dealt with by case rules. Domain independent rules are preferred, but domain specific revision rules cannot be ignored.

CASE 1: INDEPENDENCE. Many states of the world are independent of each other. Consequently the truth, falsity or certainty of some statements about the world are independent of the truth, falsity or uncertainty of others. This is trivially true when the states are unrelated to each other, but also if one believes that (P is possible) then assertions that (E is evidence against P) or (E is evidence for P) have no logical consequences for continuing to believe P is possible.

CASE 2: INCONSISTENCY. If it is asserted that (A is a conclusive argument for P) but also that (E is compelling evidence against P) then a rule of combination is inappropriate. The system should report the inconsistency (making no conclusions from its components), or assess the sources of the conflict before taking any action (essentially Cohen's approach, op cit).

CASE 3: PRECEDENCE. Suppose there is evidence or argument in favour of P, but nothing conclusive. For example there is evidence in favour of a diagnosis of peptic ulcer and against gastric cancer. Now suppose a new piece of knowledge becomes available to the effect that endoscopic investigation has revealed a clear ulcer. A conclusive state takes precedence over a probable one. Similarly a probable state takes precedence over one that is merely assumed in the absence of any information (more detail in Fox, 1985).

CASE 4: AGGREGATION. The final case is the familiar one of aggregation, where belief is a summary of evidence and argument. There is only space for one example here, so a domain specific one is taken. For example take the domain rule:

if imminent invasion of SomeCountry is ambiguous
then surveillance of SomeCountry is recommended

A proposition like 'imminent invasion of Monrovia' might be treated as ambiguous, however much evidence against it has been accrued, if the alternative is plausible. Domain knowledge might show that invasion is ambiguous because it is in the interests of some other country to invade, however much evidence there is that the country is friendly. The traditional domain independent rule that some threshold of posterior probability is exceeded is only one of a number of techniques.

## The problem of formalisation

AI methods are often informal by comparison with traditional techniques.

Sometimes the flexibility of informal knowledge may be worth the lack of formalisation. Wherever possible however we should consider the validity and limits of a method. Three criteria are considered here; completeness, consistency and correctness.

Like knowledge itself this scheme is not complete - perhaps for the same reasons that knowledge can never be complete. This paper presents three types of uncertainty, the current implementation has about a dozen which people have found to be sufficiently general that they have entered the English language (Fox, 1985). In time more may be disovered.

A formal scheme should not be inconsistent. It should not be possible to generate a contradiction from the axioms. Since the scheme is not exhaustive (complete) over some set of uncertainty axioms we cannot be sure that it will not, given some further extension, generate an inconsistency, though I have found none yet. Heuristics carry no guarantees.

In what sense could these distinctions between possibility, probability and plausibility be 'correct'? I do not know whether their correctness is formally decidable. However (1) the distinctions seem obvious; (2) ideas of possibility and plausibility have recurred in the history of probabilistic thinking, and never been entirely thrown over (Hacking, 1975); (3) they are routinely used in science and technology and similar "unknown country".

To conclude

> "I am inviting the reader to imagine ... that there is a space of possible theories about probability that has been rather constant from 1660 to the present ... perhaps an understanding of our space and its preconditions can liberate us from the cycle of probability theories that has trapped us for so long" Hacking, 1975, p 16.

Note Hacking's use of the word "possible".